# Data-driven Probabilistic Atlases Capture Whole-brain Individual Variation

Yuankai Huo[1], Katherine Swett[2], Susan M. Resnick[3], Laurie E. Cutting[2], Bennett A. Landman[1]

[1] Electrical Engineering, Vanderbilt University, Nashville, TN, USA
[2] Special Education, Vanderbilt University, Nashville, TN, USA
[3] National Institute on Aging, Baltimore, MD, United States

**Abstract.** Probabilistic atlases provide essential spatial contextual information for image interpretation, Bayesian modeling, and algorithmic processing. Such atlases are typically constructed by grouping subjects with similar demographic information. Importantly, use of the same scanner minimizes inter-group variability. However, generalizability and spatial specificity of such approaches is more limited than one might like. Inspired by Commowick's "Frankenstein's creature paradigm" which builds a personal specific anatomical atlas, we propose a data-driven framework to build a personal specific probabilistic atlas under the large-scale data scheme. The data-driven framework clusters regions with similar features using a point distribution model to learn different anatomical phenotypes. Regional structural atlases and corresponding regional probabilistic atlases are used as indices and targets in the dictionary. By indexing the dictionary, the whole brain probabilistic atlases adapt to each new subject quickly and can be used as spatial priors for visualization and processing. The novelties of this approach are (1) it provides a new perspective of generating personal specific whole brain probabilistic atlases (132 regions) under data-driven scheme across sites. (2) The framework employs the large amount of heterogeneous data (2349 images). (3) The proposed framework achieves low computational cost since only one affine registration and Pearson correlation operation are required for a new subject. Compared with site-based group atlases, the experimental results show that the proposed atlases capture more individual variations by decreasing the Jensen–Shannon divergence between probabilistic atlases and the ground truth. Our method matches individual regions better with higher Dice similarity value when testing the probabilistic atlases. Importantly, the advantage the large-scale scheme is demonstrated by the better performance of using large-scale training data (1888 images) than smaller training set (720 images).

## 1 Introduction

Probabilistic atlases play important roles in understanding the spatial variation of brain anatomy, in visualization, and in the processing of data. The basic framework of making probabilistic atlases is to bring the image data from the selected subjects into an atlas space by rigid or non-rigid registration [1]. Then, probabilistic maps are gen-erated by averaging the segmentations of regions from a specific group of subjects with similar demographic data, such as age, sex and from the same site. However, the inter-subject variability is normally larger than the inter-group variability, which causes the group-based scheme to fail to capture a great deal of individual variation.

To overcome the large inter-subject variability, Commowick et al. proposed the "Frankenstein's creature paradigm" to build a personal specific anatomical atlas for head and neck region [2]. The paradigm first selected regional anatomical atlases based on a training database then merged them together into a complete atlas. However, this framework cannot be directly applied on making probabilistic atlases since each probabilistic atlas is averaged from a group of segmentations. Moreover, compared with the 105 CT images used as the database in Commowick's framework, we employ 2349 heterogeneous MRI images in our framework.

In this paper, we propose a large-scale data-driven framework to learn a dictionary of the whole brain probabilistic atlases (132 regions) from 1888 heterogeneous 3D MRI training images. The novel contributions of this paper are (1) providing a new data-driven perspective of making whole brain probabilistic atlas, (2) generating the more accurate personal specific probabilistic atlases by using the large-scale data from different groups and even different sites, and (3) achieving low computational cost of applying the learned dictionary on new subjects.

## 2  Data

The dataset aggregates 9 datasets with a total 2349 MRI T1w 3D images obtained from healthy subjects. The 2349 images are divided to 1888 training and 431 testing datasets based on the site and demographic information. The 1888 training images are used to train the data-driven framework ("Training Set 1888"). A subset of 720 training images ("Training Set 720") is employed to generate group atlases (**Table 1**).

**Table 1.** Data summary of Training Set 720 and Testing Set

|    | Study | Site | Sex (1 is male) | Age (years) | Scanner (Tesla) | Training (number) | Testing (number) |
|----|-------|------|-----------------|-------------|-----------------|-------------------|------------------|
| 1  | BLSA | NIA | 1, 2 | 29~45 | 3T | 40 | 0 |
| 2  | Cutting | Vanderbilt | 1, 2 | 20~30 | 3T | 40 | 37 |
| 3  | ABIDE | NYU | 1, 2 | 15~32 | 3T | 40 | 0 |
| 4  | IXI | Guys | 1 | 20~45 | 1.5T | 40 | 22 |
| 5  | IXI | Guys | 2 | 20~45 | 1.5T | 40 | 20 |
| 6  | IXI | HH | 1, 2 | 20~45 | 3T | 40 | 47 |
| 7  | IXI | IOP | 1, 2 | 20~45 | 1.5T | 40 | 0 |
| 8  | ADHD200 | NYU | 1, 2 | 15~17 | 3T | 40 | 0 |
| 9  | ADHD200 | NeuroIM | 1, 2 | 15~26 | 3T | 40 | 0 |
| 10 | ADHD200 | Pittsburgh | 1, 2 | 15~20 | 3T | 40 | 0 |
| 11 | fcon_1000 | Beijing | 1 | 20~26 | 3T | 40 | 23 |
| 12 | fcon_1000 | Beijing | 2 | 20~26 | 3T | 40 | 61 |
| 13 | fcon_1000 | Cambridge | 1 | 20~25 | 3T | 40 | 17 |
| 14 | fcon_1000 | Cambridge | 2 | 21~25 | 3T | 40 | 39 |
| 15 | fcon_1000 | ICBM | 1, 2 | 19~45 | 3T | 40 | 0 |
| 16 | fcon_1000 | NewYork | 1, 2 | 20~45 | 3T | 40 | 52 |
| 17 | fcon_1000 | Oulu | 1, 2 | 20~23 | 1.5T | 40 | 63 |
| 18 | NKI_rockland | Rockland | 1, 2 | 15~45 | 3T | 40 | 35 |
|    | OASIS with manual segmentation | | 1, 2 | 18~90 | 3T | 0 | 45 |
|    | | | | | **Total** | 720 | 461 |

*The Full **Training Set 1888** is obtained from the following datasets:
BLSA: Baltimore Longitudinal Study of Aging
ABIDE: Autism Brain Imaging Data Exchange
ADHD200: Attention Deficit Hyperactivity Disorder
NKI_rockland: Nathan Kline Institute Rockland
NDAR: National Database for Autism Research

Cutting: Data from Cutting pediatric project
IXI: Information eXtraction from Images
fcon_1000: 1000 Functional Connectome
OASIS: Open Access Series on Imaging Study

## 3 Methods

The proposed data-driven framework consists of two main portions. First, a dictionary is learned by the training data (§3.1-3.3) (**Figure 1**). Second, the learned dictionary is applied to a new subject by affine alignment to MNI space (§3.4-3.5) (**Figure 2**).

### 3.1 Get Regional Segmentations and Point Distribution Model

All 720 training subjects were first affinely registered [3] to the MNI305 atlas [4]. Then, a state-of-the-art multi-atlas segmentation (including atlases selection, pairwise registration [5], label fusion [6] and error correction [7]) was performed on each subject. 45 MPRAGE images from OASIS dataset were used as original atlases which are manually labeled with 133 labels (132 brain regions and 1 background) by the Brain-COLOR protocol [8]. Here, we define $S_i$ as the whole brain segmentations with 133 labels and the $i \in \{1,2 \dots ,720\}$ represent different subjects.

Then, a mean segmentation $\bar{S}$ is generated from all $\{S_i\}_{i=1,2,\dots,720}$ by majority vote label fusion. Since the $\bar{S}$ is smooth, it is a good template of making surface meshes for 132 regions. When the meshes are generated, the vertices $\bar{V}^k$ on the mean segmentation $\bar{S}$ can be propagated to individual segmentations [9]. We non-rigidly register each $S_i$ to $\bar{S}$ and get the diffeomorphism $\phi_i(\cdot)$ [5]. The inverse transformation $\phi_i^{-1}(\cdot)$ is used to propagate the $\bar{V}^k$ back to individual vertices $V_i^k$ (Figure 1).

### 3.2 Clustering

The Affinity Propagation (AP) clustering method [10] was used to cluster the similar segmentations by using the $V_i^k$ as features. The advantage of AP clustering is it can adaptively cluster the samples into a number of clusters without providing the number of clusters. For region $k$, the negative mean Euclidian distance $d^k(i,j)$ between vertices $V_i^k$ and $V_j^k$ is used as the similarity measurement for AP clustering,

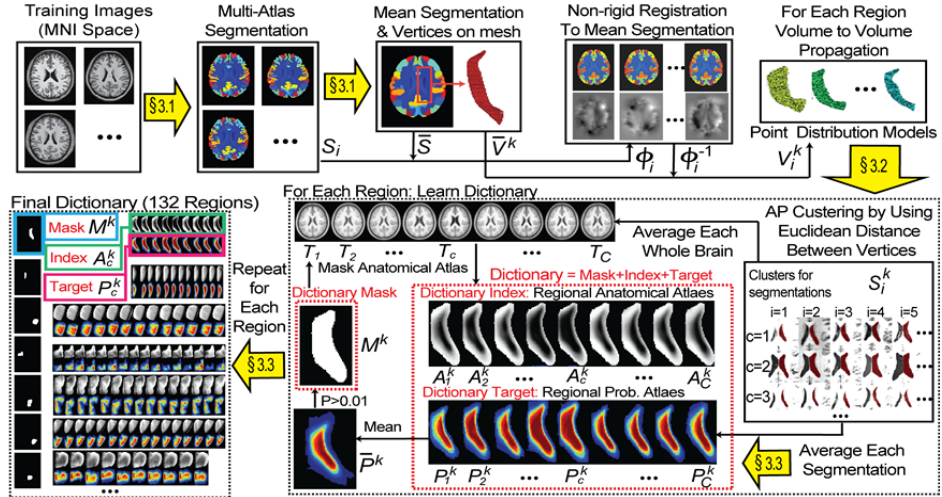

**Fig. 1.** Flowchart of training a data-driven dictionary of whole brain probabilistic atlas.

$$d^k(i,j) = -\frac{1}{M_k} \sum_{m=1}^{M_k} \left\| v_{i,m}^k - v_{j,m}^k \right\|^2 \qquad (1)$$

where the $v_{i,m}^k$ and $v_{j,m}^k$ are the $m^{th}$ vertex in the vertices $V_i^k$ and $V_j^k$. $M_k$ is the size of the vertices $V_i^k$ or $V_j^k$. Typically, 7~20 reliable clusters are generated for each region.

### 3.3 Learn Dictionary

**For One Region**

The regional anatomical atlases $A_c^k$ are the "dictionary index" and the regional probabilistic atlases $P_c^k$ corresponding "dictionary target" (red rectangular in **Figure 1**). First, the regional probabilistic atlases $P_c^k$ for the cluster $c$ is obtained by averaging the segmentations that belong to that cluster.

$$P_c^k = \frac{1}{L_c} \sum S_i^k, \quad T_c = \frac{1}{L_c} \sum I_i, \quad all\ i \in cluster\ c \qquad (2)$$

where $S_i^k$ is the segmentation of region $k$ from subject $i$ and $L_c$ is the number of segmentations in the cluster $c$. The anatomical atlases for each cluster are found by (2) and $I_i$ is the whole brain anatomical image from subject $i$.

However, as shown in **Figure 1**, each $T_c$ is a whole brain anatomical atlas rather than a regional anatomical atlas for region $k$. So, we need to extract the target area for region $k$ by a reasonable mask $M^k$.

To get the mask $M^k$, we (1) average all $\{P_c^k\}_{c=1,2,\dots,C}$ to $\bar{P}^k$ (2) obtain the $M^k$ by setting the threshold $\bar{P}^k > 0.01$. The obtained mask will be much larger than any individual segmentation, which covers the potential spatial locations of region $k$.

Finally, we apply the mask $M^k$ on every $T_c$ to get a regional anatomical atlas $A_c^k$

$$A_c^k = T_c \circ M^k \qquad (3)$$

The masked $A_c^k$ is corresponding to the regional probabilistic atlas $P_c^k$.

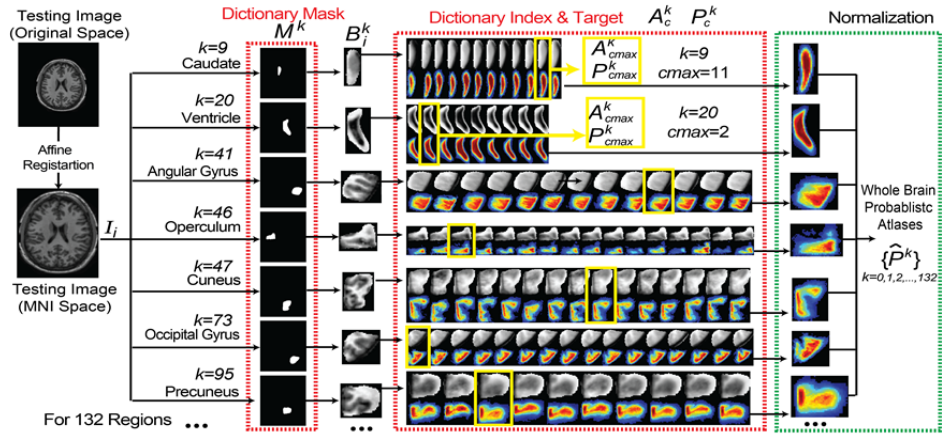

**Fig. 2.** Flowchart of applying the dictionary to customize a probabilistic atlas for a new subject.

**For Whole Brain**

We repeat the "For One Region" steps 132 times (for all regions except background) to get the whole brain dictionary as shown in the lower left part of **Figure 2**.

### 3.4 Apply Dictionary on New Subjects

To efficiently establish an individual whole brain probabilistic atlases, each target subject is affinely aligned [3] to the MNI305 atlas to get $I_i$ (**Figure 2**). Then, the regional intensity $B_i^k$ can be masked out by

$$B_i^k = I_i \circ M^k \tag{4}$$

By comparing the $B_i^k$ to our learned dictionary, the index can be obtained by finding the most correlated regional anatomical atlas $A_c^k$. The correlation metrics used here is the Pearson correlation. Once the index $c_{max}$ is found, the corresponding $P_{c_{max}}^k$ is chosen as the regional probabilistic atlas for the new subject.

$$c_{max}^k = \arg\max_c \; corr(A_c^k, B_i^k), \qquad c \in \{1,2,\dots,C\} \tag{5}$$

Repeating equations (4) and (5) for all regions, we find the 132 most correlated regional probabilistic atlases for the new subject.

### 3.5 Normalize to Whole Brain Atlas

Since the regional probabilistic atlases were chosen independently, the total probability for a voxel might be larger or smaller than 1. To normalize them to a complete set of whole brain probabilistic atlases, we employed a whole brain tissue probabilistic mask $M^t$ from 1888 training image which contains the voxels with tissue probability greater than 0.95. For each voxel $(x, y, z)$ within the mask $M^t$, the 132 regional probabilistic atlases are normalized to 1; otherwise we keep it untouched.

$$\hat{P}^k(x,y,z) = \begin{cases} \dfrac{P_{c_{max}^k}^k(x,y,z)}{Z} & x,y,z \in brain\; mask\; M^t, or\; Z > 1 \\ P_{c_{max}^k}^k(x,y,z) & otherwise \end{cases} \tag{6}$$

$Z = \sum_{k=1}^{132} P_{c_{max}^k}^k(x,y,z)$ is the normalization term.

Last, the probability of background $\hat{P}^0(x,y,z)$ is obtained by

$$\hat{P}^0(x,y,z) = 1 - \sum_{k=1}^{132} \hat{P}^k(x,y,z) \tag{7}$$

The set of $\{\hat{P}^k(x,y,z)\}_{k=0,1,2,\dots,132}$ is the normalized data-driven whole brain probabilistic atlases for the new subject. For each voxel in the whole brain probabilistic atlases, the total probability of 132 labels and background is 1.

## 4 Experimental Results

Two metrics are employed in the experiments. First, the Jensen-Shannon (JS) divergence is used to assess the spatial similarity between the probabilistic atlases and the target segmentations for each testing subject [11]. Here, the "target segmentations" means the multi-atlas segregations for the withheld testing images and the manual segmentations for the OASIS images. The smaller JS divergence value is, the more similar the two spatial distributions are. So, smaller is the better for JS.

Second, to compare the different probabilistic atlases more intuitively, we apply "naive segmentation" on whole brain by choosing labels with the highest probability for each voxel. Notice that we are not providing a novel segmentation algorithm. Instead, we compare the spatial accuracy of different probabilistic atlases by using the naïve segmentation since this approach is entirely depending on the probability. Then, the Dice similarity measures the overlaps between the naive segmentations and the target segmentations.

All statistical significance tests are made using a Wilcoxon signed rank test ($p<0.01$). Creating a whole brain probabilistic atlas for a new subject can be done with 1 rigid registration and 12 seconds of CPU time (Xeon W3520 2.67GHz).

### 4.1 Evaluation by Withheld Testing Data

**Figures 3** and **4** show the results using withheld testing subjects. The green boxplots represent the average JS or Dice values by applying the probabilistic atlases from all the other 17 group atlases for one testing subject. The blue, red and orange boxplots show the JS or Dice values by using the corresponding group probabilistic atlases, data-driven probabilistic atlases from Training Set 720 and from Training Set 1888.

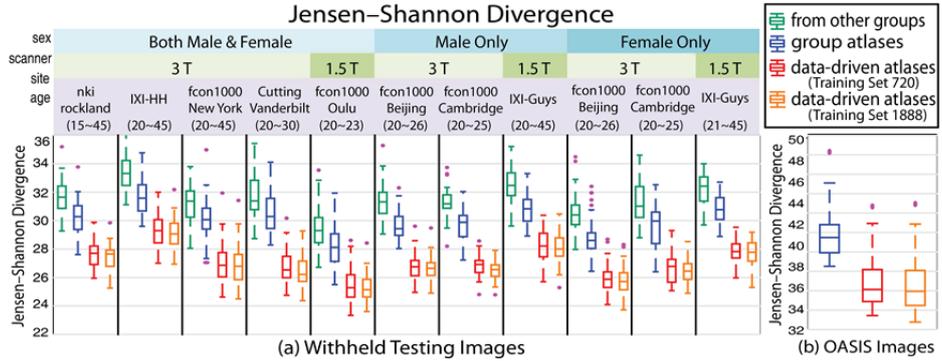

**Fig. 3.** Jensen-Shannon divergence. The comparisons of JS divergence for different atlases are all significantly different for both withheld and OASIS testing images.

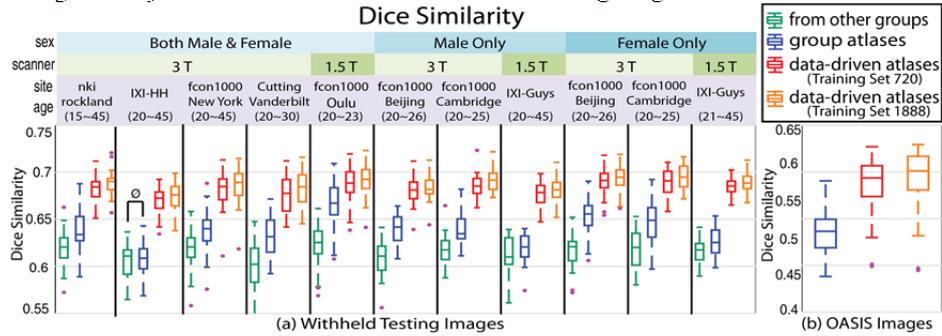

**Fig. 4.** Dice similarity. The comparisons of Dice value for different atlases are all significant for both withheld and OASIS testing images except the IXI-HH group marked by "Ø".

**Figure 3** and **4** demonstrate that the data-driven atlases match the target segmentations significantly better than the traditional group based atlases with the significantly smallest JS divergence and greatest Dice values while the atlases from other groups perform the worst. Moreover, for the data-driven atlases with two different numbers of training images, the large-scale Training Set 1888 performs significant better than Training Set 720 for both JS divergence and Dice similarities.

To conclude, (1) the group based atlases perform significantly better than the atlases from other groups which demonstrates the group-based framework is able to control the inter-group variability; (2) our proposed data-driven framework produced the more accurate probabilistic atlases than group based atlases by capturing the individual variance; (3) by using the large-scale training data, the performance of data-driven framework is improved significantly.

### 4.2 Evaluation by OASIS Data

45 subjects from OASIS dataset with manual segmentations are used for 44 leave one tests. The data-driven probabilistic atlases are obtained from the learned dictionary. The right hand panel of results in **Figures 3 and 4** show that the results of manual segmentations repeat the finding in §4.1.

Moreover, we show one testing subject (slice z = 75 in MNI space from 3D image) from the OASIS dataset in **Figure 5**. By comparing with the manual segmentations for 6 regions, it shows that the data-driven atlases match the true segmentations more accurately than the group atlases. Moreover, the large-scale Training Set 1888 matches the manual segmentation better than the smaller Training Set 720.

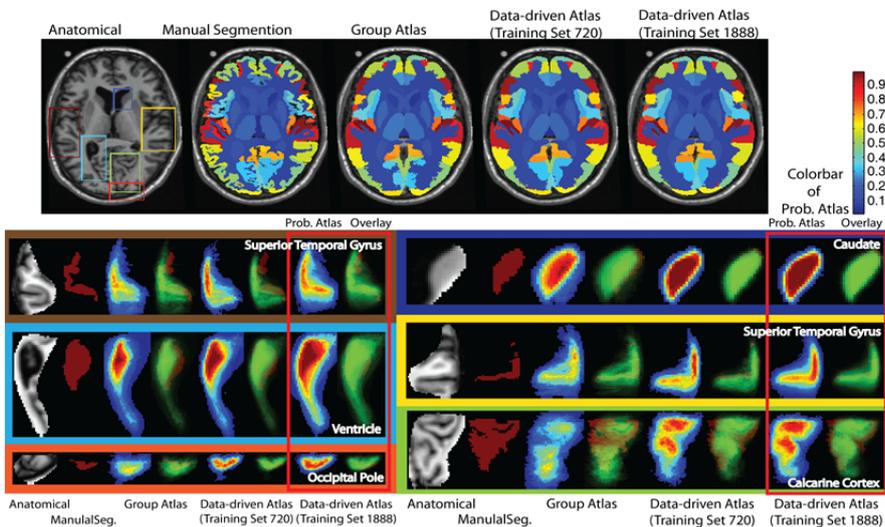

**Fig. 5.** One testing subject from OASIS dataset. Top row shows the anatomical image, manual segmentation, highest probability segmentations using the group probabilistic atlases, Training Set 720 and Training Set 1888. The lower rows show the details of 6 regions. For each region, from left to right are: anatomical image, manual segmentation, probabilistic atlases generated by different methods and their overlays on manual segmentations.

## 5  Discussion

We present a data-driven framework to learn a dictionary of whole brain probabilistic atlases to achieve accurate individualized whole brain probabilistic atlases. This framework (1) provides a new perspective of using data-driven scheme rather than the traditional group based methods, (2) uses the large-scale heterogeneous data to achieve more personal specific probabilistic atlases than using the single-group and single-site data by capturing the individual variation (3) demonstrates the advantages of using large-scale scheme in generating personal probabilistic atlases compared with the smaller size of training images, and (4) only requires one affine registration and Pearson correlations to apply to new subjects which achieves low computational cost.

Due to the higher accuracy and low computational cost, the proposed method is able to be the priors in many medical image processing algorithms and applications.

**Acknowledgments:** This research was supported by NIH 5R21EY024036, NIH 1R21NS064534, NIH 2R01EB006136, NIH 1R03EB012461, NIH R01EB006193 and also supported by the Intramural Research Program, National Institute on Aging, NIH.

## References

1. Shattuck, D.W., et al.: Construction of a 3D probabilistic atlas of human cortical structures. NeuroImage 39, 1064-1080 (2008)
2. Commowick, O., et al.: Using Frankenstein's creature paradigm to build a patient specific atlas. Medical image computing and computer-assisted intervention : MICCAI ... International Conference on Medical Image Computing and Computer-Assisted Intervention 12, 993-1000 (2009)
3. Ourselin, S., et al.: Reconstructing a 3D structure from serial histological sections. Image Vision Comput 19, 25-31 (2001)
4. Evans, A.C., et al.: 3D statistical neuroanatomical models from 305 MRI volumes. In: Nuclear Science Symposium and Medical Imaging Conference, 1993., 1993 IEEE Conference Record., pp. 1813-1817. IEEE (1993)
5. Avants, B.B., et al.: Symmetric diffeomorphic image registration with cross-correlation: evaluating automated labeling of elderly and neurodegenerative brain. Medical image analysis 12, 26-41 (2008)
6. Asman, A.J., Landman, B.A.: Non-local statistical label fusion for multi-atlas segmentation. Medical image analysis 17, 194-208 (2013)
7. Wang, H., et al.: A learning-based wrapper method to correct systematic errors in automatic image segmentation: consistently improved performance in hippocampus, cortex and brain segmentation. NeuroImage 55, 968-985 (2011)
8. Klein, A., et al.: Open labels: online feedback for a public resource of manually labeled brain images. In: 16th Annual Meeting for the Organization of Human Brain Mapping. (2010)
9. Heimann, T., Meinzer, H.P.: Statistical shape models for 3D medical image segmentation: a review. Medical image analysis 13, 543-563 (2009)
10. Frey, B.J., Dueck, D.: Clustering by passing messages between data points. Science 315, 972-976 (2007)
11. Gouttard, S., et al.: Assessment of reliability of multi-site neuroimaging via traveling phantom study. Medical image computing and computer-assisted intervention : MICCAI ... International Conference on Medical Image Computing and Computer-Assisted Intervention 11, 263-270 (2008)